\documentclass[conference,onecolumn]{IEEEtran}
\IEEEoverridecommandlockouts
\usepackage{cite}
\usepackage{amsmath,amssymb,amsfonts}
\usepackage{graphicx}
\usepackage{textcomp}
\usepackage{xcolor}
\usepackage{listings}
\def\BibTeX{{\rm B\kern-.05em{\sc i\kern-.025em b}\kern-.08em
    T\kern-.1667em\lower.7ex\hbox{E}\kern-.125emX}}
\begin{document}

\title{Automatic Differentiation for Inverse Problems with Applications in Quantum Transport\\
}

\author{\IEEEauthorblockN{1\textsuperscript{st} Ivan Williams}
\IEEEauthorblockA{\textit{Electrical and Computer Engineering} \\
\textit{University of Massachusetts Amherst}\\
Amherst, United States \\
inwilliams@umass.edu}
\and
\IEEEauthorblockN{2\textsuperscript{nd} Eric Polizzi}
\IEEEauthorblockA{\textit{Electrical and Computer Engineering, Mathematics}\\
\textit{University of Massachusetts Amherst}\\
Amherst, United States \\
epolizzi@engin.umass.edu}
}

\maketitle

\begin{abstract}
A neural solver and differentiable simulation of the quantum transmitting boundary model is presented for the inverse quantum transport problem. The neural solver is used to engineer continuous transmission properties and the differentiable simulation is used to engineer current-voltage characteristics.
\end{abstract}

% Note that keywords are not normally used for peerreview papers.
\begin{IEEEkeywords}
QTBM, PINN, AD, SciANN, JAX
\end{IEEEkeywords}

\IEEEpeerreviewmaketitle

\section{Introduction}

\IEEEPARstart{T}{he} modern engineering design process is largely guided by computational models. Devices and systems are often first prototyped 'in-silico'. While standard numerical techniques, such as finite-element, have served well in the forward case, other computational tasks, such as model inversion, remain challenging.

Physics-informed machine learning is a mesh-free method for solving differential equations. By embedding the governing equations and boundary conditions of a physical system into the loss function of a neural network, the network can be trained to approximate the solution to the system to a high accuracy \cite{raissi2017physics}.

The physics-informed approach allows the solving of a physical model from a symbolic specification. This enables two important applications for computational modeling in engineering; solving models in parameter space and solving inverse and ill-posed problems. 

The aforementioned advantages make the physics-informed approach well suited for problems in quantum transport. In the quasi-static approximation of quantum transport, the system energy is taken as a parameter. Using physics-informed neural networks, the solution to the quantum transmitting boundary problem is solved for every energy in a given range simultaneously allowing for the direct evaluation of transmission properties. Further, the internal potential can be represented as a neural network as well, and it can be trained to reproduce specified transmission properties (this work). 

Despite the advantages of the physics-informed approach. These techniques are often affected %plagued
by problems of spectral bias and stiffness \cite{chuang2023predictive}. Steep gradients in the true solution confound the NN and require very fine-step size and/or long training times. This is why physics-informed approaches often incorporate data and require significant computational resources.   

Underpinning the physics-informed approach is automatic differentiation (AD) which allows for the optimization of the network under the constraints of the physical system. It's possible to combine the advantages of numerical linear algebra with the physics-informed approach utilizing AD tools directly \cite{rackauckas2021universal}. Algorithms specifying matrix computations and other numerical techniques can be differentiated with respect to their parameters thus algorithms specifying approximate physical solutions can be differentiated with respect to parameters of the physical system.

The differentiable approach is well suited to inverse problems. In quantum transport, this can be exploited to engineer an internal barrier to satisfy particular current-voltages characteristics as it will be discussed in this work.

% You must have at least 2 lines in the paragraph with the drop letter
% (should never be an issue)

\section{Model}
We utilize the quantum transmitting boundary model \cite{Lent1990,Polizzi2000}. Charge is injected into a 1D quantum wire from the left.

For the the physics-informed neural network the model is represented as two coupled ODEs corresponding to the real and imaginary components of the quantum wavefunction. The continuity equation is included as well; PINNs benefit from conservation law regularizations \cite{wang2022modified}. In the inverse problem, the internal potential is also a neural network. The target is encoded as a boundary condition along the domain edge. 

For the differentiable simulation the wavefunction is computed on a fixed grid and transmission and current is computed with a fixed discretization. The barrier is modeled by two box functions parametrized by their position, width and height. The current-voltage characteristics are optimized with respect to the barrier parameters and the Fermi level.

\subsection{Physics-Informed}
The wavefunction is separated into its real and imaginary parts as follows:

\begin{equation}
  \Psi = \Psi_{R} + i \Psi_{I}.
\end{equation}

The wavenumbers corresponding to the source $k_1$ and drain $k_2$ of the device are given by:

\begin{equation}
    k_1 = \sqrt{\frac{2m}{\hbar^2}(E-V(x=0))},
\end{equation}

\begin{equation}
    k_2 = \sqrt{\frac{2m}{\hbar^2}(E-V(x=L))}.
\end{equation}

The following are the boundary conditions in the quasi-static approximation with current being injected from the left:

\begin{equation}
    \left. \frac{d\Psi_{R}(x; E)}{dx} \right|_{x=0} = k_1 \Psi_{I}(0; E),
\end{equation}

\begin{equation}
    \left. \frac{d\Psi_{R}(x; E)}{dx} \right|_{x=L} = -k_2 \Psi_{I}(L; E),
\end{equation}

\begin{equation}
    \left. \frac{d\Psi_{I}(x; E)}{dx} \right|_{x=0} = 2k_1 - k_1 \Psi_{R}(0; E),
\end{equation}

\begin{equation}
    \left. \frac{d\Psi_{I}(x; E)}{dx} \right|_{x=L} = k_1 \Psi_{R}(L; E).
\end{equation}

The governing equation in quantum transport is the Time-Independent Schr\"odinger Equation:

\begin{equation}
    \left[ -\frac{\hbar^2}{2m} \frac{d^2}{dx^2} + V(x) \right] \Psi_{R}(x; E) = E \Psi_{R}(x; E),
\end{equation}

\begin{equation}
     \left[ -\frac{\hbar^2}{2m} \frac{d^2}{dx^2} + V(x) \right] \Psi_{I}(x; E) = E \Psi_{I}(x; E). 
\end{equation}
Where $V(0\leq x\leq L)$ is an arbitrary total internal potential including the barrier potential (in quantum transport, $V(x\leq 0)=0$ and $V(x\geq L)=V_{0}$).

The following constraint enforces the conservation of probability:

\begin{equation}
    \frac{d}{dx} \left( \Psi_{R} \frac{d\Psi_{I}}{dx} - \Psi_{I} \frac{d\Psi_{R}}{dx} \right) = 0.
\end{equation}

Lastly, for the inverse problem, the additional constraint enforces the transmission properties are a desired target relation, $T(E)$:

\begin{equation}\label{TE}
  \frac{k_2}{k_1}|\Psi(x=L; E)|^2 = T(E).
\end{equation}

\subsection{Differentiable}
Parametrized by height, center and width; $\vec{\theta} = (H, C, W)$,
we define the barrier function, $B(x; \vec{\theta})$, with convenience functions $\text{LHS} (x;\vec{\theta})$ and $\text{RHS} (x; \vec{\theta})$ to model internal barriers as follows:
\begin{equation}
\text{LHS} (x;\vec{\theta} = (H, C, W)) = H\tanh\left(x-\frac{L(2C-W)}{2}\right),
\end{equation}

\begin{equation}
\text{RHS} (x;\vec{\theta} = (H, C, W)) = H\tanh\left(x-\frac{L(2C+W)}{2}\right),
\end{equation}

\begin{equation}
    B(x; \vec{\theta}) = \frac{\text{LHS} (x; \vec{\theta}) - \text{RHS} (x; \vec{\theta})}{2}.
\end{equation}

The total internal parametrized potential, $U(x; \vec{\phi}, E, V_0)$, is given by:
\begin{equation}
U(x; \vec{\phi}, E, V_0) = B_{DBL}(x; \vec{\theta}_1, \vec{\theta}_2) - V_{BIAS}(x; V_0) - E,
\end{equation}
where $B_{DBL}(x; \vec{\phi})$ and $V_{BIAS}(x; V_0)$ are the internal barrier and bias across the device, respectively:
\begin{equation}
B_{DBL}(x; \vec{\phi} = (\vec{\theta}_1, \vec{\theta}_2)) = B(x; \vec{\theta}_1) + B(x; \vec{\theta}_2),
\end{equation}
\begin{equation}
V_{BIAS}(x; V_0)= V_0\frac{x}{L}.
\end{equation}

After discretization, the Schr\"odinger equation for quantum transport is modelled by the following linear system:
\begin{equation}\label{ls}
(\mathcal{H}+\mathcal{U}+\mathcal{B})\mathbf{\vec{\Psi}} = \mathbf{\vec{S}}.
\end{equation}
 Using a 1D finite different method (FDM), $\mathcal{H}$ becomes the following Hamiltonian matrix:
\[
\begin{bmatrix}
    \alpha  & -\alpha &        &        &        \\
    -\alpha & 2\alpha   & -\alpha    &        &        \\
        & -\alpha & 2\alpha  & -\alpha    &        \\
        &     & \ddots & \ddots & \ddots \\
        &     &        & -\alpha & \alpha  
\end{bmatrix}
\]
with $\alpha=\hbar^2/(2m_e*a_n^2)$, $a_n=L/(n-1)$ ($m_e$ is the electron mass, $\hbar$ is the 
Planck constant, $L$ is the length of the 1D device and $n$ is the number of FDM dicretization points).

In (\ref{ls}), $\mathcal{U}$ is the total internal potential (diagonal matrix after discretization):
\[
\begin{bmatrix}
U(x_1; \vec{\phi}, E, V_0)&&&\\
&U(x_2; \vec{\phi}, E, V_0)&&\\
&&\ddots&\\
&&&U(x_n; \vec{\phi}, E, V_0)
\end{bmatrix}
\]
and $\mathcal{B}$ represents the quantum transmitting boundary conditions:
\[
\begin{bmatrix}
    - \text{BND}_1(E, V_0) & & & &\\
     & 0 & & &\\
     & & \ddots & &\\
      & & & 0 &\\
     & & & &- \text{BND}_2(E, V_0)
\end{bmatrix}
\]
where $\text{BND}_j(E, V_0)$ corresponds to the boundary conditions at the $j$th terminal of the 1D device:
\begin{equation}
\text{BND}_j(E, V_0) = ik_j(E, V_0) \alpha a_n. 
\end{equation}

Finally, $\mathbf{\vec{S}}$ is the source term corresponding to the charge injected at $x=0$:
\[
\begin{bmatrix}
    2\text{BND}_1(E, V_0) \\
    0 \\
    0 \\
    \vdots \\
    0
\end{bmatrix}
\]
and $\mathbf{\vec{\Psi}}$ is the discretized wavefunction vector in (\ref{ls}). %approximate wavefunction. %; $\mathbf{\vec{\Psi}(\vec{x})} \approx \Psi(x)$.

Using $T(E; V_0, \vec{\phi})$ defined in (\ref{TE}), the quantum %classical
current through the device is given as follows (at 0K) :
\begin{equation}
    I(V_0; \vec{\Phi} = (\vec{\phi}, \mu)) =\int_{\mu-V_0}^{\mu} T(E; V_0, \vec{\phi}) \, dE, 
\end{equation}
where $\mu$ is the Fermi level (given) and $\vec{\Phi}$ is a vector of all parameters of interest; internal barrier parameters ($\vec{\phi}$ $\equiv$ ($\vec{\theta}_1$, $\vec{\theta}_2)$) and the Fermi level.

Given $m$ desired current voltage measurements (observations), $\vec{O}_m$, the error, $\vec{L}(\vec{\Phi}; \vec{O}_m)$, and Loss, $\vec{\mathcal{L}}(\vec{\Phi}; \vec{O}_m)$, are given by:
\begin{equation}
    \vec{V}_{T} = (V_{T1}, V_{T2}, V_{T3}, ..., V_{Tm}),
\end{equation}

\begin{equation}
    \vec{I}_{T} = (I_{T1}, I_{T2}, I_{T3}, ..., I_{Tm}),
\end{equation}

\begin{equation}
    \vec{L}\Big(\vec{\Phi}; \vec{O}_m \equiv (\vec{I}_{T}, \vec{V}_{T})\Big) = I(\vec{V}_{T}; \vec{\Phi}) - \vec{I}_{T},
\end{equation}

\begin{equation}
    \vec{\mathcal{L}}(\vec{\Phi}; \vec{O}_m) \equiv \frac{\vec{L}(\vec{\Phi}; \vec{O}_m) \cdot \vec{L}(\vec{\Phi}; \vec{O}_m) }{m}.
\end{equation}

The Inverse Problem is formulated as the following optimization problem:
\begin{equation}
    \underset{\vec{\Phi}}{\text{minimize}} \quad \mathcal{J}(\vec{\mathcal{L}}(\vec{\Phi}; \vec{O}_m)), % \\ 
\end{equation}
where $\mathcal{J}(\vec{\mathcal{L}}(\vec{\Phi}; \vec{O}_m))$ is the Jacobian of the Loss.

\section{Results}
\subsection{Physics-Informed}
In the physics-informed case, both the wavefunction, $\Psi(x; E)$, and total internal potential, $V(x)$, are neural networks that co-optimized in the training process to reproduce a target transmission curve. The bias $V_0$ is taken as -0.375eV, with the internal potential fixed at each terminal; $V(x=0) = 0$ and $V(x=L) = V_0$. The network is trained on Fourier features, that is, in the Fourier domain, to suppress the effect of spectral bias. The API, SciANN \cite{Haghighat_2021}, is used for all neural network training.

\begin{figure}[ht]
    \centering
    \includegraphics[width=0.4\textwidth]{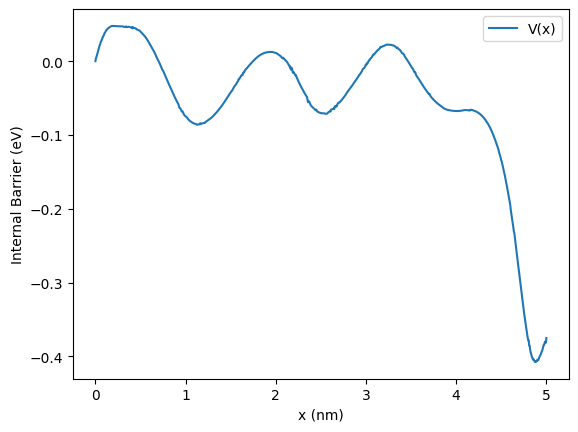}
    \caption{\label{fig1} Engineered internal potential.}
\end{figure}

\begin{figure}[ht]
    \centering
    \includegraphics[width=0.4\textwidth]{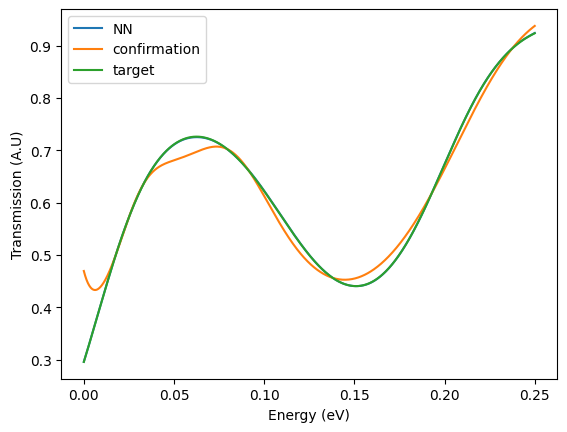}
    \caption{\label{fig2} Target transmission properties (green) vs. the neural network prediction (blue) and the 'confirmation' transmission properties (orange).}
\end{figure}

\begin{figure}[ht]
    \centering
    \includegraphics[width=0.4\textwidth]{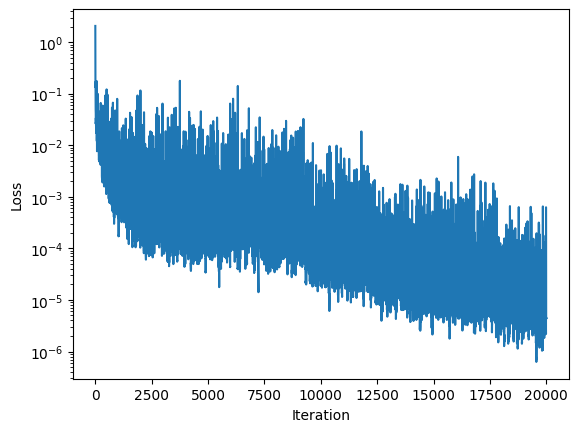}
    \caption{\label{fig3} History of model loss over 20,000 training iterations.}
\end{figure}

Fig. \ref{fig1} and \ref{fig2} show the optimized barrier and corresponding transmission curves, respectively. The internal potential is optimized
as a neural network to satisfy the desired targeted quantum transmission curve. The history of the Loss is shown in Fig. \ref{fig3}. The orange curve in Fig \ref{fig2} is the "confirmation" transmission curve obtained when the optimized potential in Fig \ref{fig1}, is passed to a finite-difference solver.

\subsection{Differentiable}
For the differentiable simulation, differentiation is implemented with JAX \cite{jax2018github} and optimization is implemented with Optax \cite{deepmind2020jax}. JAX is employed due to it's autovectorization functionality; $\mathbf{vmap()}$. Computation of transmission properties and current-voltage characteristics require evaluations of continuous functions of parameters, thus it is necessary to map the linear solve over $E$ to compute the transmission properties and to map the computation of the transmission over $V$ to compute the current.

\lstset{language=Python, 
        basicstyle=\ttfamily\small, 
        keywordstyle=\color{blue}, 
        stringstyle=\color{green},
        commentstyle=\color{red},
        showstringspaces=false,
        %numbers=left,
        %numberstyle=\tiny\color{gray},
        captionpos=b}

The Loss is computed by the functions below.

\begin{lstlisting}
def TSM(E,V,theta,theta2): return k2(E,V)*\
(abs(QT(E,V,theta,theta2))**2)[N-1]/\
(k1(E)+ 1E-10)

def I(V,theta,theta2,uf): return np.trapz(\
np.interp(np.linspace(uf-V,uf,M2),\
np.linspace(0,uf,M),\
vmap(TSM,(0,None,None,None))\
(np.linspace(0,uf,M),V,theta,theta2)),\
np.linspace(uf-V,uf,M2))

def Loss(P): return\
((vmap(I,(0,None,None,None))\
(vT,P[0:3],P[3:6],P[6]) - iT)**2).mean()

J = jacfwd(Loss)
\end{lstlisting}

JAX employs operator overloading to differentiate through the functions $\tt Loss({P})$, $\tt {I}({V},{theta},{theta2},{uf})$ and $\tt {TSM}({E},{V},{theta},{theta2})$.

The internal potential is optimize via the training loop below where the Adabelief optimizer is utilized for its combination of adaptive learning and performance \cite{zhuang2020adabelief}:
\begin{lstlisting}
optimizer = \
optax.adabelief(learning_rate=0.001)

state = optimizer.init(initial_params)
for i in range(num_iterations):
    gradient = vmap(J)(initial_params)
    updates, state = optimizer.\
    update(gradient, state, initial_params)
    updated_params = optax.\
    apply_updates(initial_params, updates)
    initial_params = updated_params
\end{lstlisting}

\begin{figure}[ht]
    \centering
    \includegraphics[width=0.4\textwidth]{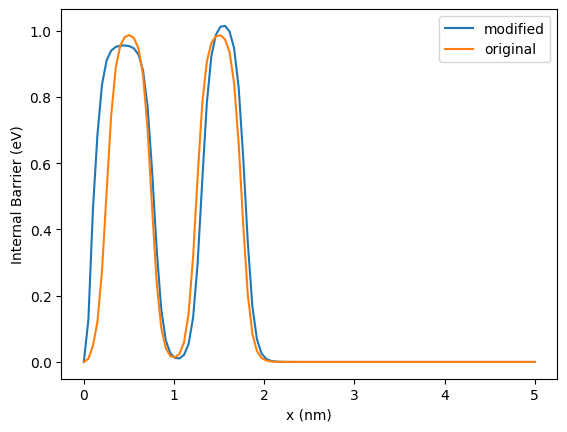}
    \caption{Original potential (orange) vs. modified potential (blue).}
\end{figure}

\begin{figure}[ht]
    \centering
    \includegraphics[width=0.4\textwidth]{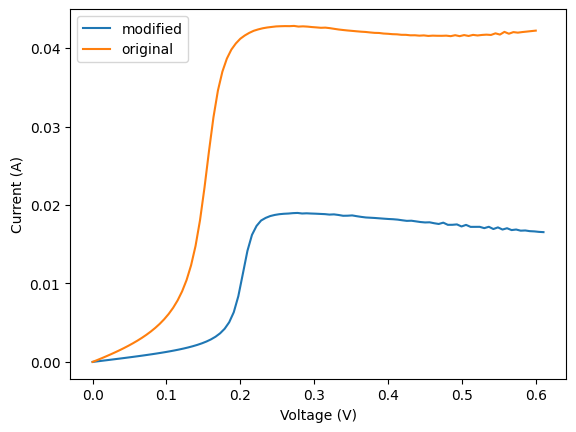}
    \caption{Original current-voltage (orange) vs. target current-voltage (blue).}
\end{figure}

Fig. 4 and 5, show the internal double barrier and current-voltage characteristics of two resonant-tunneling type devices. The 'modified' device is specified to dissipate less power and have a greater negative differential resistance than the 'original', demonstrating a plausible application in the design of low-power amplifiers \cite{lee2014reflection}. The double barrier is contained within a window function to penalize barrier configurations inside the terminals; $U(x=0)-U(x=L) \neq V_0$ is unphysical. 

%\section{Discussion}
%Absolutus Obseletum

\section{Conclusion}
 A neural solver and differentiable simulation are demonstrated for inverse quantum transport. The neural solver can engineer continuous electron-wave transmission properties and the differentiable simulation can engineer current voltage relations.

Future work will focus on the inverse problem in the context of differentiable simulation due to flexibility of AD tools compatible with existing numerical algorithm paradigms. Extension to finite-element and generalizing the internal potential to a neural network is a natural next step. Incorporation of self-consistent methods are necessary to model devices in the many-body regime. Other possible applications may include multi-dimensional electron-wave devices \cite{Polizzi2002}.
\appendices
\section{Table of Hyperparameters}

\begin{table}[htbp]
\centering
\begin{tabular}{|c|c|}
\hline
PINNs & parameters  \\
\hline
NN size (all) & [8X20] \\
\hline
n & 1000 \\
\hline
FF(samples,std.dev.,mean) & 20,1,0 \\
\hline
Train(step,iterations,samples) & 0.001,20000,20000 \\
\hline
\end{tabular}
%\caption{Parameters for the PINNs model.}
\end{table}

\begin{table}[htbp]
\centering
\begin{tabular}[htbp]{|c|c|}
\hline
DS & parameters  \\
\hline
FD discretization & 100 \\
\hline
Energy Points & 100 \\
\hline
Interpolation Points & 100 \\
\hline
m & 2\\
\hline
Search Space & 25\\
\hline
optimizer iterations & 1000\\
\hline
\end{tabular}
%\caption{Parameters for the DS model.}
\end{table}
\bibliographystyle{ieeetr}
\bibliography{RQE}

\end{document}